\documentclass[letterpaper]{article} % DO NOT CHANGE THIS
\usepackage{aaai24}  % DO NOT CHANGE THIS
\usepackage{times}  % DO NOT CHANGE THIS
\usepackage{helvet}  % DO NOT CHANGE THIS
\usepackage{courier}  % DO NOT CHANGE THIS
\usepackage[hyphens]{url}  % DO NOT CHANGE THIS
\usepackage{graphicx} % DO NOT CHANGE THIS
\usepackage{natbib}  % DO NOT CHANGE THIS AND DO NOT ADD ANY OPTIONS TO IT
\usepackage{caption} % DO NOT CHANGE THIS AND DO NOT ADD ANY OPTIONS TO IT
\usepackage{algorithm}
\usepackage{algorithmic}

\usepackage[dvipsnames]{xcolor}  % https://en.wikibooks.org/wiki/LaTeX/Colors
\usepackage{colortbl}
\usepackage{amsmath}
\usepackage{amssymb}
\usepackage{booktabs}
\usepackage{multirow}
\usepackage{bm}
\usepackage{xspace}
\usepackage{arydshln}
\usepackage{lipsum}
\usepackage{txfonts}
\usepackage{indentfirst}
\usepackage{newfloat}
\usepackage{listings}
\DeclareCaptionStyle{ruled}{labelfont=normalfont,labelsep=colon,strut=off} % DO NOT CHANGE THIS
\floatstyle{ruled}
\newfloat{listing}{tb}{lst}{}
\floatname{listing}{Listing}
\title{Expressive Forecasting of 3D Whole-body Human Motions}
\author{
    Pengxiang Ding\textsuperscript{\rm 1,\rm 2},\,\,\,  Qiongjie Cui\textsuperscript{\rm 3,\rm 5}\thanks{Corresponding author}
    \\ 
    Min Zhang\textsuperscript{\rm 1},\,\,
    Mengyuan Liu\textsuperscript{\rm 4},\,\,
    Haofan Wang\textsuperscript{\rm 5},\,\, 
    Donglin Wang\textsuperscript{\rm 1}
}
\affiliations{
    \textsuperscript{\rm 1}MiLAB, Westlake University,\,\,\,
    \textsuperscript{\rm 2}Zhejiang University,\,\,\,   
    \textsuperscript{\rm 3}Nanjing University of Science and Technology\\ 
    \textsuperscript{\rm 4}Shenzhen Graduate School, Peking University,\,\,\,
    \textsuperscript{\rm 5}Xiaohongshu Inc.
    \\
    
    dingpengxiang@westlake.edu.cn, \,\,
    cuiqiongjie@njust.edu.cn
}

\usepackage{bibentry}
\begin{document}

\maketitle

\begin{abstract}
Human motion forecasting, with the goal of estimating future human behavior over a period of time, is a fundamental task in many real-world applications.
However, existing works typically concentrate on predicting the major joints of the human body without considering the delicate movements of the human hands.
In practical applications, hand gesture plays an important role in human communication with the real world, and expresses the primary intention of human beings.
In this work, we are the first to formulate a whole-body human pose forecasting task, which jointly predicts the future body and hand activities. Correspondingly, we propose a novel Encoding-Alignment-Interaction (EAI) framework that aims to predict both coarse (body joints) and fine-grained (gestures) activities collaboratively, enabling expressive and cross-facilitated forecasting of 3D whole-body human motions.
Specifically, our model involves two key constituents: cross-context alignment (XCA) and cross-context interaction (XCI).
Considering the heterogeneous information within the whole-body, XCA aims to align the latent features of various human components, while XCI focuses on effectively capturing the 
context interaction among the human components. 
We conduct extensive experiments on a newly-introduced large-scale benchmark and achieve state-of-the-art performance.  The code is public for research purposes at https://github.com/Dingpx/EAI.
\end{abstract}

\section{Introduction} \label{sec:introduction}
Predicting the evolution of human behavior/activity in the physical world over time is an essential aspect of machine intelligence~\cite{tarvainen2017mean,Ruiz2018HumanMP,yuan2020dlow}.
For instance, to make the seamless human-robot interaction (HRI), a robot is supposed to have some notion of how people will move or act in the near future, conditioned on a series of historically observed poses~\cite{gui2018adversarial,Cui_2021_CVPR,Zhang_2021_CVPR,mao2019learning,cai2020learning,Dang_2021_ICCV,ding2021uncertainty}.

\begin{figure}[t]
\centering
  \includegraphics[width=3.5in]{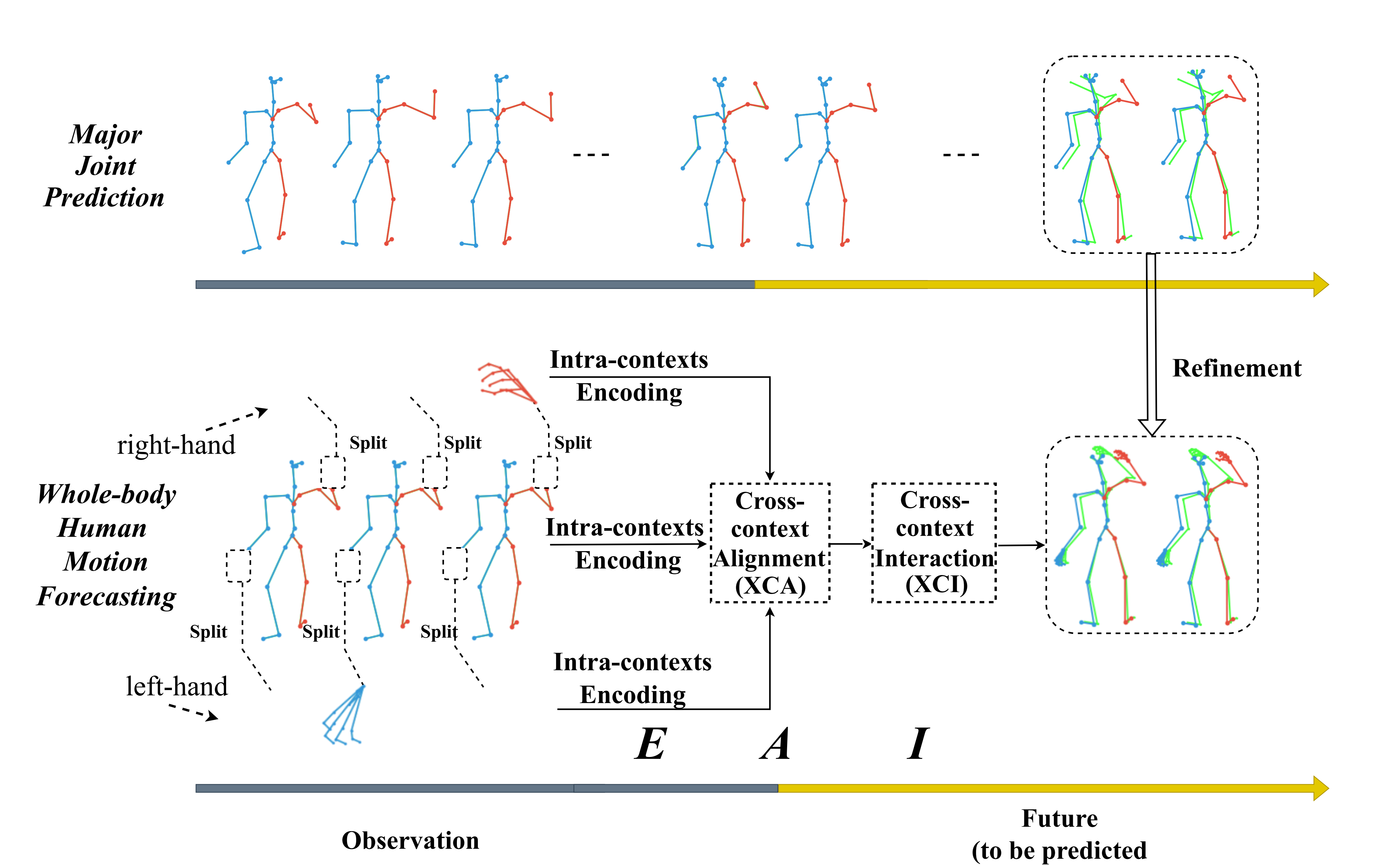} 
\vspace{-1.5em}
\caption{
Top: Previous works focus on predicting the human major joints, without considering delicate hand movements that are critical to the HRI application.   
Bottom: To fill this gap, our work proposes a novel task: whole-body human pose forecasting, to jointly predict future both body and gesture activities.
We also highlight that within the proposed EAI, both coarse- (major joints) and fine-grained (gestures) properties are cross-facilitated to achieve a higher-fidelity prediction.
Here, red/blue pose is the predicted result, while the underlying green is the ground truth.
}
\vspace{-1em}
\label{Fig:teaser}
\end{figure}

Over the past few years, this attractive topic has received considerable attention, emerging a large number of approaches, with deep learning techniques proving to be sought-after~\cite{cai2021unified,feng2021collaborative,li2022skeleton,petrovich2021action,Ruiz2018HumanMP,vaswani2017attention}. 
Moreover, we note that existing works fall into the coarse-grained scope, \textit{i.e.}, forecasting major joint movements of the human body~\cite{Adeli_2021_ICCV,Cui_2021_CVPR,butepage2017deep,Ruiz2018HumanMP,zhong2022spatio,ma2022progressively}. 
However, in terms of realistic applications, it remains a significant limitation: the subtle activity (\textit{i.e.}, gestures) is not considered. 
The human hand is a vital bridge for interacting with the world, and meanwhile, for the HRI application, it typically includes a detailed command to the robot, embodying human behaviors with the major body~\cite{Zhang_2021_CVPR,diller2022forecasting,jin2020whole,hidalgo2019single,taheri2020grab}.
% In addition, for certain particular populations (\textit{e.g.}, lower limb disabilities), the gestures may even be the sole medium for robots to understand their intents \cite{guo2021human,routray2021hand,van2011combining,yang2019gesture}.
From real applications of human pose forecasting, we, therefore, suggest that considering only major joints, while ignoring the subtle hand gestures, is not sufficient.

To fully investigate this issue, we propose a novel paradigm: \textit{whole-body human motion forecasting}, that is, conjointly predicting future activities of all joints within the body and hands, as shown in Figure \ref{Fig:teaser}.
In contrast to the conventional task, it presents significant challenges in the following aspects:
{1)} %It has the 
{There are} distinct motion patterns %of
{in} major body and gesture (amplitude of movement, skeletal freedom), and hence it is sub-optimal to treat them equally;
{2)} %For various activities, it involves
{A human activity usually involves} %obvious
collaboration/interactivity of different parts within the whole-body;
For instance, the clapping-hand embodies the interaction of both hands; and for drinking, {it is} %its 
dominated by the semantic association of the hands and mouth.
{3)} Due to the heterogeneous scales and characteristics, it is not feasible to directly model such cross-grained interaction {as} existing multi-person interactive forecasting methods do \cite{guo2022multi}.

In this work, we propose a novel Encoding-Alignment-Interaction (EAI) framework to address these challenging issues.
Specifically, to avoid negative mutual-interference, we first extract their separate internal spatio-temporal correlations from the body and gesture's heterogeneous motion properties.
We observe that, the interaction/collaboration of various elements within the whole-body is critical %for to 
{for}
performing a specific activity.
However, such interaction is incompatible with the existing multi-person interaction \cite{guo2022multi}, because person-to-person information is scale-uniform, whereas intra-body context is heterogeneous,  \textit{e.g.}, coarse-to-fine-grained (body-to-gesture), or vice versa.
Therefore, we propose to exploit the cross-context alignment (XCA) to effectively align and smooth the latent features of different parts, thus eliminating their heterogeneity. 
Finally, with the aligned features, we further introduce cross-context interaction (XCI), a variant of cross-attention \cite{hao2017end}, that is able to capture the pairwise interactivity between various human parts within the whole-body.
We note that, the proposed EAI is a generic framework capable of simutaneously consider the interactivity of different parts within the whole-body as well as the heterogeneous properties, resulting in the higher-quality whole-body prediction.

Our contributions are as follows:
(1) {To the best of our knowledge, this work is the first to} predict the future actions of major joints and human gestures simultaneously.
(2) We propose a Encoding-Alignment-Interaction (EAI) approach, equipped with the XCA and XCI, which is capable of extracting the heterogeneous interaction within the whole body.
(3) Extensive experiments show {that} our model achieves superior performance for both short- and long-term prediction compared to the competitors.

\section{Related Work} \label{sec:relatedwork}
\newcommand{\highlightSYMBOLS}[1]{\xspace{\color{black} #1}\xspace}
\newcommand{\hpose}{\textbf{X}}
\newcommand{\vhpose}{\textbf{x}}
\newcommand{\fpose}{\textbf{Y}}
\newcommand{\adjmatrix}{\textbf{A}}
\newcommand{\weimatrix}{\textbf{W}}
\newcommand{\intrafeatures}{\textbf{S}}
\newcommand{\SPoutput}{\mathbf{\widetilde S}}
\newcommand{\dctmatrix}{\textbf{C}}
\newcommand{\SpatialEncoding}{\textbf{E}}
\newcommand{\Fusionoutput}{{\textbf{F}}}
\newcommand{\Fusionoutputupdate}{\widetilde {{\textbf{F}}}}
\newcommand{\query}{\textbf{Q}}
\newcommand{\key}{\textbf{K}}
\newcommand{\valu}{\textbf{V}}
\newcommand{\amap}{\textbf{M}}
\newcommand{\weightm}{{\textbf{{W}}}}
\newcommand{\historytime}{{T}}
\newcommand{\futuretime}{{\Delta T}}
\newcommand{\wholebodyskeletonumber}{{ N}}
\newcommand{\intraencoderweightmatrixnumber}{{ F}}
\newcommand{\intraencodernumber}{{ {N_{1}}}} 
\newcommand{\intraencoderweightmatrixdimension}{{ H}}
\newcommand{\dctnumber}{{ H_{c}}}
\newcommand{\mblhDNweights}{{ \alpha}}
\newcommand{\wholebodycordinatenumber}{{ D}}
\newcommand{\xcinumber}{{ {N_{2}}}} 
\newcommand{\lh}{l}
\newcommand{\rh}{r}
\newcommand{\mb}{m}
\newcommand{\intraencoderlayernumbers}{n}
\newcommand{\Expection}{\operatorname{{E}}}
\newcommand{\Variance}{\operatorname{{Var}}}
\newcommand{\hposes}{{\hpose}}
\newcommand{\fposes}{{\fpose}}
\newcommand{\lhhposes}{{\hpose}_{{\lh}}}
\newcommand{\lhfposes}{{\fpose}_{{\lh}}}
\newcommand{\mbhposes}{{\hpose}_{{\mb}}}
\newcommand{\mbfposes}{{\fpose}_{{\mb}}}
\newcommand{\rhhposes}{{\hpose}_{{\rh}}}
\newcommand{\rhfposes}{{\fpose}_{{\rh}}}
\newcommand{\mbdctposes}{{\hpose}^{''}_{{\mb}}}
\newcommand{\lhpfposes}{{\hat{{\fpose}}_{\lh}}}
\newcommand{\mbpfposes}{{\hat{{\fpose}}_{\mb}}}
\newcommand{\rhpfposes}{{\hat{{\fpose}}_{\rh}}}
\newcommand{\rawmapping}{\mathcal{M}}
\newcommand{\newmapping}{\mathcal{M}_{\textsc{WB}}}
\newcommand{\intramodule}{Intra-context Encoding}
\newcommand{\intramodulemini}{intra-context encoding }
\newcommand{\lhintrafeatures}{{{\intrafeatures}}_{\lh}}
\newcommand{\mbintrafeatures}{{{\intrafeatures}}_{\mb}}
\newcommand{\rhintrafeatures}{{{\intrafeatures}}_{\rh}}
\newcommand{\lhskeletonumber}{{\wholebodyskeletonumber_{\lh}}}
\newcommand{\rhskeletonumber}{{\wholebodyskeletonumber_{\rh}}}
\newcommand{\mbskeletonumber}{{\wholebodyskeletonumber_{\mb}}}
\newcommand{\lhadjacencymatrix}{\adjmatrix_{\lh}}
\newcommand{\mbadjacencymatrix}{\adjmatrix_{\mb}}
\newcommand{\lhintrafeaturescom}{{{\intrafeatures}}_{\lh}{i}}
\newcommand{\mbintrafeaturescom}{{{\intrafeatures}}_{\mb}^{i}}
\newcommand{\rhintrafeaturescom}{{{\intrafeatures}}_{\rh}^{i}}
\newcommand{\lhweightmatrix}{\weimatrix_{\lh}}
\newcommand{\mbweightmatrix}{\weimatrix_{\mb}}
\newcommand{\intraencoderactf}{\sigma}
\newcommand{\crossalignmodule}{Cross-context Alignment (XCA)}
\newcommand{\crossalignmodulemini}{cross-context alignment}
\newcommand{\distribution}{{q}}
\newcommand{\mbdistribution}{{\distribution}_{\mb}}
\newcommand{\lhdistribution}{{\distribution}_{\lh}}
\newcommand{\DNLayer}{\operatorname {CN}}
\newcommand{\DNoutput}{\intrafeatures}
\newcommand{\lhDNoutput}{{{\DNoutput}}_{{\lh}}^{'}}
\newcommand{\lhDNoutputend}{{\widetilde \DNoutput}_{\lh}}
\newcommand{\mbDNoutput}{{{\DNoutput}}_{{\mb}}^{'}}
\newcommand{\mbDNoutputend}{{\widetilde \DNoutput}_{\mb}}
\newcommand{\rhDNoutput}{{{\DNoutput}}_{{\rh}}^{'}}
\newcommand{\rhDNoutputend}{{\widetilde \DNoutput}_{\rh}}
\newcommand{\mbrhDNweights}{\beta}
\newcommand{\rhlhDNweights}{\gamma}
\newcommand{\SPLayer}{\operatorname {SFP}}
\newcommand{\lhSPoutput}{{\SPoutput}_{{\lh}}}
\newcommand{\mbSPoutput}{{\SPoutput}_{{\mb}}}
\newcommand{\rhSPoutput}{{\SPoutput}_{{\rh}}}
\newcommand{\MMD}{\operatorname {MMD}}
\newcommand{\Avg}{\operatorname {Avg}}
\newcommand{\LossMMD}{\mathcal{L}^{dis}}
\newcommand{\lhmbMMDLoss}{\LossMMD_{{\lh}{\mb}}}
\newcommand{\mbrhMMDLoss}{\LossMMD_{{\mb}{\rh}}}
\newcommand{\rhlhMMDLoss}{\LossMMD_{{\rh}{\lh}}}
\newcommand{\crossinteractionmodule}{Cross-context Interaction (XCI)}
\newcommand{\crossinteractionmodulemini}{cross-context interaction}
\newcommand{\ModeratorLayer}{\operatorname{MLP}}
\newcommand{\tempweight}{\tau}
\newcommand{\lhMODinput}{{\Fusionoutput}_{{\lh}}^{lw}}
\newcommand{\mbMODinput}{{\Fusionoutput}_{{\mb}}^{lw}}
\newcommand{\rhMODinput}{{V}_{{\rh},{rwrist}}}
\newcommand{\lhMODoutput}{{\Fusionoutputupdate}_{{\mb}}^{lw}}
\newcommand{\rhMODoutput}{{\Fusionoutputupdate}_{{\mb}}^{rw}}
\newcommand{\lhmbMODweights}{w_{{\lh}{\mb}}}
\newcommand{\rhmbMODweights}{w_{{\rh}{\mb}}}
\newcommand{\SpatialEncodingFunction}{\operatorname{PT}}
\newcommand{\etal}{\mbox{et al.}\xspace}
\newcommand{\ie}{\mbox{i.e.}\xspace}
\newcommand{\eg}{\mbox{e.g.}\xspace}
\newcommand{\e}[1]{\ensuremath{\times 10^{#1}}}

% \textbf{Stochastic Human Motion Forecasting.}
% Considering the diversity and stochasticity, researchers leverage the generative model ( \textit{\eg}, VAEs, GANs)~\cite{yan2018mt,ZhouLXHH018,yuan2020dlow,mao2021generating} to achieve the probabilistic prediction. 
% The stochastic approach generates multiple potential actions; however, in real-world HRI, robots can only execute one specific instruction, making deterministic methods essential.

\textbf{Human Motion Forecasting.}
%recurrent-based
RNNs are the widely-used architectures for time-series data modeling and human pose prediction \cite{butepage2017deep,butepage2018anticipating,honda2020rnn,corona2020context}.
Despite encouraging progress, they typically suffer from error accumulation and tend to converge to a static pose. 
Feed-forward networks, such as convolutional neural networks (CNNs) \cite{liu2020trajectorycnn,ding2022towards} and graph neural networks (GNNs)\cite{mao2019learning,mao2020history,li2021rain,li2022skeleton}, are proposed as an alternative solution to alleviate the drawbacks of recurrent models. 
\cite{mao2019learning} presents the learnable adjacent matrix to model spatial dependencies among human body joints. 
This approach is later extended with self-attention on an entire piece of historical information~\cite{mao2020history} or a selection of them~\cite{li2021rain}. 
Despite the promising performance, the existing approaches all fall into the scope of predicting the motions of major human joints, without co-analyzing the hand gestures \cite{pavlakos2019expressive,Cui_2021_CVPR}.
From the realistic application, we note that subtle hand movements are indispensable for the expressive human behavior and intention.
Our work first notices this challenging issue, \textit{\ie}, how to predict the expressive whole-body human motion (unifying the gesture and major human joints), and aims to solve it.

\textbf{Contextual Interaction.}
Contextual Interaction have proven to be effective in the human-to-human interactions~\cite{guo2022multi,wang2021multi,rong2021frankmocap}. 
Specifically, \cite{wang2021multi} models the context of individual motion and social interactions through a Multi-Range Transformers structure.
\cite{guo2022multi} explores the multi-person contextual interactions via a designed cross-interaction module. However, the interaction/collaboration of various components within the whole-body is incompatible with the above method because human-to-human information is scale-uniform, whereas the intra-body context is heterogeneous. Therefore, in our proposed EAI, we introduce the alignment of heterogeneous features across human components to extract subsequent whole-body internal interactivity more effectively.

\begin{figure*}[!t]
\centering
  \includegraphics[width=.95\textwidth]{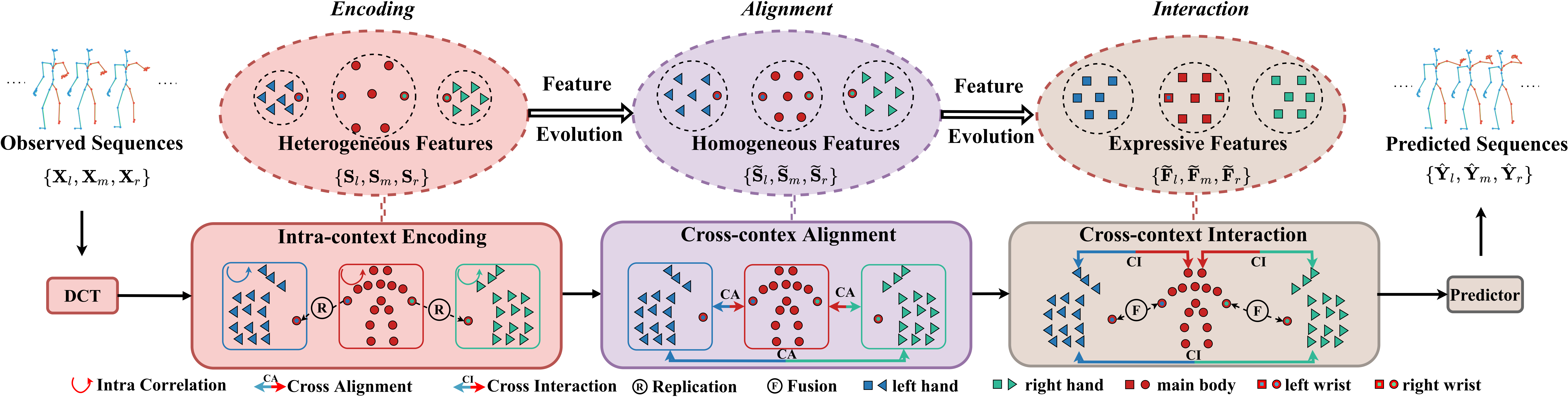}
  \vspace{-.5em}
\caption{
Overall framework of encoding-alignment-interaction (EAI).
Given the observed whole-body sequences $\{{\mathbf{X}_{l},\mathbf{X}_{m},\mathbf{X}_{r}}\}$,
we first achieve the heterogeneous features 
$\{{\mathbf{ S}_{l},\mathbf{S}_{m},\mathbf{ S}_{r}}\}$
via intra-context encoding for each body component independently. 
Since those intra-context lacks the interaction information of components, the cross-context alignment (XCA) and the cross-context interaction (XCI) are the subsequently proposed to extract cross-context information, where the former aims to alleviate the heterogeneity of components to generate homogeneous features while the latter is designed to explore the cross-context interaction according to the homogeneous features $\{{\mathbf{\widetilde S}_{l},\mathbf{\widetilde S}_{m},\mathbf{\widetilde S}_{r}}\}$ from the XCI. The resulting expressive features
$\{{\widetilde{\mathbf{F}}_{l},\widetilde{\mathbf{F}}_{m},\widetilde{\mathbf{F}}_{r}}\}$
are then used to predict future whole-body sequences $\{{\hat{\mathbf{Y}}_{l},\hat{\mathbf{Y}}_{m},\hat{\mathbf{Y}}_{r}}\}$.
}
\vspace{-1em}
\label{Fig:framework}
\end{figure*}

\vspace{-.5em}
\section{Proposed Method} \label{sec:method}
\vspace{-.2em}
\textbf{Problem Setup.}
Previous works typically consider the forecasting of the major human joints.
Given ${\historytime}$ history human poses ${\hposes}=[\vhpose_1,\vhpose_2,...,\vhpose_{\historytime}]$,
informally, its objective can be defined as learning a mapping ${\rawmapping}: {\hposes} \xrightarrow{} {\fposes}$ to estimate the future poses ${\fposes}$, where ${\hposes}$ is the observed major joints, ${\fposes}=[\vhpose_{{\historytime}+1},\vhpose_{{\historytime}+2},...,\vhpose_{{\historytime}+\futuretime}]$ is the corresponding future ones over ${\futuretime}$ frames.
This work extends the above standard setup to united whole-body human motion forecasting, including major body, left and right hand, denoted by $m$, $l$, $r$ variables for the sake of simplicity.
Analogously, we define the novel task as learning a united mapping $\mathcal{M}_{\text{WB}}$:
\begin{equation}
    \resizebox{0.7\linewidth}{!}{${\newmapping}: {\{ {\lhhposes}, {\mbhposes}, {\rhhposes}\}}
   \xrightarrow{} 
   {\{ {\lhfposes},{\mbfposes}, {\rhfposes}\}}\;,$}
\end{equation}

\noindent{where ${\mbhposes} \in \mathbb{R}^{ {\wholebodycordinatenumber}_{\mb} \times {\historytime} }$ 
(${\mbfposes} \in \mathbb{R}^{ {\wholebodycordinatenumber}_{\mb} \times {\futuretime} }$) 
is the past (future) skeletal sequence of major body. ${\wholebodycordinatenumber}_{\mb} = 3{\wholebodyskeletonumber}_{\mb}$ is the number of 3D joint coordinates in a single frame and ${\wholebodyskeletonumber}_{\mb}$ is the number of major body joints. Also, ${\lhhposes}\;({\rhhposes})$ and ${\lhfposes}\;({\rhfposes})$ are the past and future motion of the left (right) hands.
}

\subsection{{\intramodule}}
\label{subsec:intra}
Due to the distinct motion patterns of major body and gestures, we should consider the different body parts individually.
Notably, we extract the intra-context of 3D skeletal sequences comprising of left hand, head, and right hand positions, denoted as $\{{\lhhposes},\ {\mbhposes},\ {\rhhposes}\}$, in feature space. This is because the spatio-temporal correlations in feature space are more expressive than the correlations in the original motion space. Next, we illustrate the details of the encoding process by taking the major body as a specific example.

In the temporal domain, Discrete Cosine Transform (DCT) is exploited to capture the temporal smoothness by transforming the observed sequence into trajectory space. Given the past motion ${\mbhposes}$, we compute the DCT coefficients of this sequence ${\mbdctposes} \in \mathbb{R}^{\wholebodycordinatenumber_{\mb} \times {\dctnumber}}$ as:
\begin{equation}
    \resizebox{0.23\linewidth}{!}{${\mbdctposes} = {{{\hpose}^{'}_{{\mb}}}}{\dctmatrix}\;,$}
\end{equation}

\noindent where ${{\hpose}^{'}_{{\mb}}}\in \mathbb{R}^{\wholebodycordinatenumber_{\mb} \times ({\historytime+\futuretime})}$ is a variant of ${\mbhposes}$ by replicating the last observed pose $\futuretime$ times following \cite{mao2019learning}; ${\dctmatrix}\in\mathbb{R}^{(\historytime+\futuretime)\times \dctnumber}$ is the predefined DCT matrix and each row of ${\dctmatrix}$ is the DCT coefficients for a trajectory. 

In the spatial domain, we exploit GCNs~\cite{mao2019learning,Cui_2021_CVPR} to denote the skeleton as a fully-connected graph, depicted as an adjacency matrix $ {\mbadjacencymatrix} \in \mathbb{R} ^{\wholebodycordinatenumber_{\mb} \times \wholebodycordinatenumber_{\mb}}$.
Formally, we define $ {\intrafeatures}^{(n)}_{m} \in \mathbb{R} ^{\wholebodycordinatenumber_{\mb} \times {\intraencoderweightmatrixnumber}^{(\intraencoderlayernumbers)}} $ as the input feature of $\intraencoderlayernumbers$-th layer in GCNs, and
$ {\weimatrix}^{(n)}_{m} \in \mathbb{R} ^{ {\intraencoderweightmatrixnumber}^{(\intraencoderlayernumbers)} \times {\intraencoderweightmatrixnumber}^{(\intraencoderlayernumbers+1)} } $ 
as the weight matrix. Then the output feature ${\intrafeatures}_{m}^{(n+1)}$ is derived as:
\begin{equation}
    \resizebox{0.43\linewidth}{!}{${\intrafeatures}_{m}^{(n+1)}={\intraencoderactf}({\adjmatrix}_{\mb}^{(n)}{\intrafeatures}_{m}^{(n)}{\weimatrix}_{m}^{(n)})\;,$}
\end{equation}
where ${\intrafeatures}_{m}^{(1)} = {\mbdctposes}$ is the input feature and ${\intraencoderweightmatrixnumber}^{(1)} = \dctnumber$ at the first layer; The number of hidden layers $\intraencoderweightmatrixnumber^{(n)}$ are set to $\intraencoderweightmatrixdimension_{d}$; {$ {\intraencoderactf}(\cdot) $ is an activation function}. 
The final output features of last layer are ${\intrafeatures}_{m}^{(last)} \in \mathbb{R}^{ {\wholebodycordinatenumber}_{\mb} \times \intraencoderweightmatrixdimension}$, w.r.t  $\mathbf{S}_{m}$. 

Following the above formalism, we also attain the intra-context of the left (right) hands, forming the whole-body intra-context $\{ {\lhintrafeatures},\ {\mbintrafeatures},\ {\rhintrafeatures}\}$. 
We note that, although in the standard anatomy the wrist is considered to come from the major body, due to the physical connection with hands, we also include it in the hands feature extraction.
Accordingly, the feature dimension of the hands is slightly changed as 
${\lhintrafeatures} \in \mathbb{R}^{ ({\wholebodycordinatenumber}_{\lh}+3) \times \intraencoderweightmatrixdimension }$ 
and 
$ {\rhintrafeatures}\in \mathbb{R}^{ ({\wholebodycordinatenumber}_{\rh}+3) \times \intraencoderweightmatrixdimension}$. 

In light of the heterogeneity and interactivity across part-specific representations, we therefore propose cross-context alignment (XCA) and cross-context interaction (XCI) to effectively capture the cooperation within the whole-body in the following sections.

\vspace{-.1em}
\subsection{\crossalignmodule}
\vspace{-.1em}
\label{subsec:XCA}
In contrast to person$\leftrightarrow$person interactivity~\cite{guo2022multi} task, the intra-body context of different body component is scale-inconsistent due to the distinctive motion patterns.
The intra-body context is heterogeneous, coarse-to-fine-grained (body-to-gesture), or vice versa, in contrast to the current multi-person interaction where person-to-person information is scale-uniform.
In other words, these differences in motion patterns need to be alleviated because they may enormously disturb the overall motion perception. 

Intuitively, the heterogeneous context within the whole-body mainly originated from the amplitude of movement, scale, and skeletal freedom of different parts, typically reflecting the discrepancy of feature distribution~\cite{li2018adaptive}.
To address this issue, we introduce cross-neutralization among the body components combined with discrepancy constraints to effectively align the latent features of different parts. Specifically, we neutralize the distribution discrepancy among different features via a learnable factor that can be automatically adjusted according to the MMD constraint. This reorganizes the original features into new features with a closer distribution.
Such a strategy is able to alleviate the incompatibility of different body components while being more conducive to extracting the interaction.
We take the alignment process of the major body and left hand as an example.

\begin{figure}[!t]
\begin{center}
  \includegraphics[width=0.43 \textwidth]{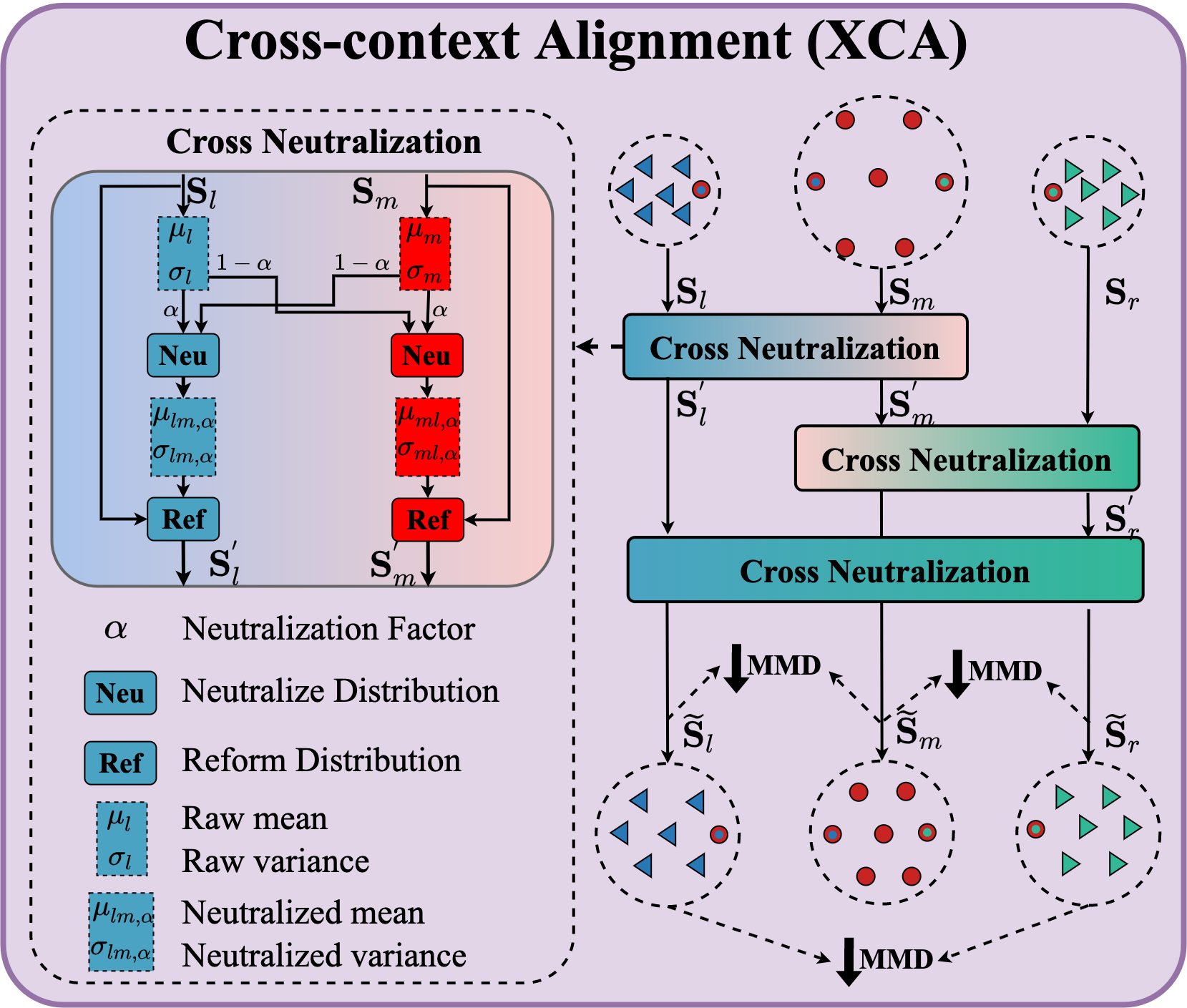} %插入图片，[]中设置图片大小，{}中是图片文件
\end{center}
\vspace{-1em}
\caption{Based on $\{ {\lhintrafeatures},{\mbintrafeatures},{\rhintrafeatures}\}$, XCA applies circular cross neutralization and discrepancy constraint (MMD) to alleviate the heterogeneity across components and generate the homogeneous features.
}
\vspace{-1em}
\label{Fig:XCA}
\end{figure}

\noindent{\textbf{Cross Neutralization (CN). }}Given the intra-context $ \{ {\lhintrafeatures}$, ${\mbintrafeatures} \}$, we introduce a learnable factor ${\mblhDNweights} \in[0.5,1]$ to constitute the fused distribution to neutralize the distributions discrepancy between ${\lhintrafeatures}$ and ${\mbintrafeatures}$. Formally, the $\DNLayer(\cdot)$ is defined as:
\begin{equation}
\begin{aligned}
 &{\bm{\mu}}_{lm,\alpha} = \alpha \bm{\mu}_{l} + (1-\alpha)\bm{\mu}_{m},{\bm{\sigma}}_{lm,\alpha} = \alpha \bm{\sigma}_{l} + (1-\alpha)\bm{\sigma}_{m}, \\
 &{\bm{\mu}}_{ml,\alpha} = \alpha \bm{\mu}_{m} + (1-\alpha)\bm{\mu}_{l},{\bm{\sigma}}_{ml,\alpha} = \alpha \bm{\sigma}_{m} + (1-\alpha)\bm{\sigma}_{l}, \\
&\quad \quad \quad \textbf{s}^{\prime}_l = \frac{\textbf{s}_l-{\bm{\mu}}_{{\lh}{\mb},{\mblhDNweights}}}{\sqrt{\epsilon+{\bm{\sigma}}_{{\lh}{\mb},{\mblhDNweights}}^2}},\textbf{s}^{\prime}_m = \frac{\textbf{s}_m-{\bm{\mu}}_{ml,{\mblhDNweights}}}{\sqrt{\epsilon+{\bm{\sigma}}_{ml,{\mblhDNweights}}^2}},
\end{aligned}
\end{equation}

\noindent where $\bm{\mu}_{l}=\operatorname{Avg}({\lhintrafeatures})$  and $\bm{\sigma}_{l}=\operatorname{Var}({\lhintrafeatures})$ are the mean and variance of intra-context features ${\lhintrafeatures}$;
$\bm{\mu}_{m}$ and $\bm{\sigma}_{m}$ can be obtained similarly;
$\operatorname{Avg}(\cdot)$ and $\operatorname{Var}(\cdot)$ is the operation to calculate mean and variance along the joint dimension;  $\bm{\mu}_{l},\bm{\mu}_{m},\bm{\sigma}_{l}, \bm{\sigma}_{m} \in \mathbb{R}^{ \intraencoderweightmatrixdimension}$; 
$\bm{\mu}_{lm,\alpha}$ ($\bm{\sigma}_{lm,\alpha}$) $\in \mathbb{R}^{ \intraencoderweightmatrixdimension}$ are the mean and variance vector of fused features distribution; 
$\textbf{s}_m$ ($\textbf{s}^{}_l$), $\textbf{s}^{\prime}_m$ ($\textbf{s}^{\prime}_l$)  $\in \mathbb{R}^{ \intraencoderweightmatrixdimension}$ are the row vector of intra-context features $\textbf{S}_m$ ($\textbf{S}^{}_l$) and fused features $\textbf{S}^{\prime}_m \in \mathbb{R}^{{\wholebodycordinatenumber}_{\mb} \times \intraencoderweightmatrixdimension}$ ($\textbf{S}^{\prime}_l\in \mathbb{R}^{({\wholebodycordinatenumber}_{\lh}+3) \times \intraencoderweightmatrixdimension}$);
$\epsilon =\text{e}^{-5}$ is a factor to avoid numerical issues.
To further part-to-part alignment, we extend $\DNLayer(\cdot)$ to the circular version:
\begin{equation}
\begin{aligned}
&     {\lhDNoutput},{\mbDNoutput} = \DNLayer({\lhintrafeatures},{\mbintrafeatures},{\mblhDNweights}),\; {\mbDNoutputend},{\rhDNoutput} = \DNLayer({\mbDNoutput},{\rhintrafeatures},{\mbrhDNweights}),\\
&\qquad \qquad \qquad 
{\rhDNoutputend},{\lhDNoutputend} = \DNLayer({\rhDNoutput},{\lhDNoutput},{\rhlhDNweights}),
\end{aligned}
\end{equation}

\noindent where
${\mbDNoutput}\; ({\mbDNoutputend}) \in \mathbb{R}^{{\wholebodycordinatenumber}_{\mb} \times \intraencoderweightmatrixdimension}$,
${\lhDNoutput}\; ({\lhDNoutputend}) \in \mathbb{R}^{({\wholebodycordinatenumber}_{\lh}+3) \times \intraencoderweightmatrixdimension}$, 
and ${\rhDNoutput}\;({\rhDNoutputend}) \in \mathbb{R}^{({\wholebodycordinatenumber}_{\rh}+3) \times \intraencoderweightmatrixdimension}$ 
are the intermediate (output) features of the circular $\DNLayer(\cdot)$; ${\mbrhDNweights} $ and ${\rhlhDNweights} $ are the factors similar to ${\mblhDNweights}$, which are updated in the training phase.

\noindent{\textbf{Discrepancy Constraint. }} We apply maximum mean discrepancy (MMD) to alleviate the part-to-part discrepancy.
\begin{equation}
\begin{aligned}
&{\lhmbMMDLoss}= {\MMD}({\Avg}({\lhDNoutputend}),{\Avg}({\mbDNoutputend})),\\
\end{aligned}
\end{equation}

\noindent where ${\Avg(\cdot)}$ is the average operation along the spatial dimension, and ${\Avg}({\lhDNoutputend})$/${\Avg}({\mbDNoutputend}))
\in \mathbb{R}^{1 \times \intraencoderweightmatrixdimension}$. ${\mbrhMMDLoss}$ and ${\rhlhMMDLoss}$ can be obtained similarly.

With the cross neutralization and discrepancy constraint, where the distribution discrepancy of intra-context features is reduced, our proposed cross-context alignment (XCA) could alleviate the heterogeneity of different intra-context. Next, we present cross-context interaction (XCI) to explore the interactions within the whole body that provides the vital cues to perceive future human intention.

\subsection{{\crossinteractionmodule}}
\label{subsec:XCI}

\begin{figure}[!t]
\begin{center}
  \includegraphics[width=0.41 \textwidth]{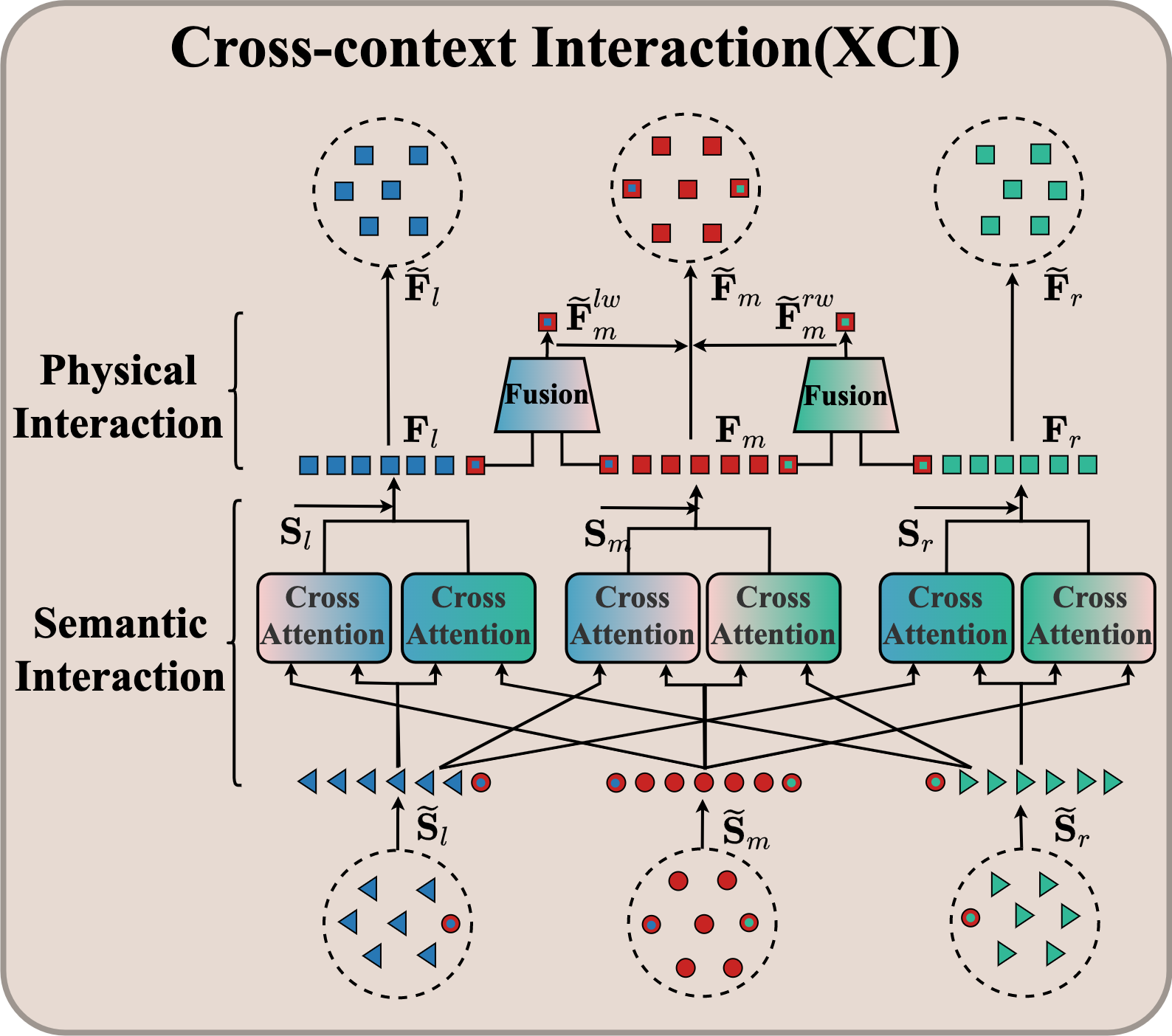} 
\end{center}
\vspace{-1em}
\caption{Taking $\{ {\lhSPoutput}, {\mbSPoutput}, {\rhSPoutput} \}$ as the input, the XCI explores the pairwise interactivity of different parts from both the semantic and physical interaction within the whole-body.}
\vspace{-1em}
\label{Fig:XCI}
\end{figure}

In contrast to the person$\leftrightarrow$person external interaction ~\cite{guo2022multi}, the body$\leftrightarrow$hands/hands$\leftrightarrow$hands involves the internal interaction across different parts within the whole body.
To be precise, it includes both semantic interaction (driven by the collaboration of different parts to perform a specific action) and
physical interaction (inherent in the chain link via the body$\leftrightarrow$hands wrist). 
Thus, we present a variant of cross-attention \cite{hao2017end} to capture the semantic and physical interactivity of various human parts.

\textbf{Semantic Interaction. }
The relevance of the three body parts is mainly derived from the mutual semantic interaction within the integrated action. 
For example, for the action of eating, fingers and head joints have strong correlations. Therefore, we aim to model the semantic dependency across components via cross-attention mechanism~\cite{hao2017end}. We take the cross-context semantic interaction between the major body and left hand as an example. The whole process is described as:
\begin{equation}
\begin{aligned}
    & {\Fusionoutput}_{{\lh}{\mb}}^{(1)} = {\SPoutput}_{{\mb}}, {\query}_{\mb}^{(n)} = {\Fusionoutput}_{{\lh}{\mb}}^{(n)} {\weightm}_{m}^{(n)} \\
    & {\key}_{\lh}^{(n)} = {\SPoutput}_{{\lh}}{\weightm}_{\lh}^{(n)},{\valu}_{\lh}^{(n)} = {\SPoutput}_{{\lh}}{\weightm}_{{\lh}{'}}^{(n)},\\
    & {\amap}_{att,{\lh}{\mb}}^{(n)} =  \operatorname{Softmax}({\query}_{\mb}^{(n)} {{\key}_{\lh}^{(n)}}^\mathrm{T}),\\
    &{\Fusionoutput}_{{\lh}{\mb}}^{(n+1)}={\Fusionoutput}_{{\lh}{\mb}}^{(n)} + \operatorname{FFN}({\amap}_{att,{\lh}{\mb}}^{(n)}{\valu}_{\lh}^{(n)}),
\end{aligned}
\end{equation}
where ${\Fusionoutput}_{{\lh}{\mb}}^{(n)}$ (${\Fusionoutput}_{{\lh}{\mb}}^{(n+1)}$) is the input (output) features; 
The input feature of the first layer is ${\SPoutput}_{{\mb}} \in \mathbb{R} ^{{{{\wholebodycordinatenumber}_{\mb}}} \times \intraencoderweightmatrixdimension}$ 
and the output features of the last layer ${\Fusionoutput}_{{\lh}{\mb}}^{(last)} \in \mathbb{R} ^{{{\wholebodycordinatenumber}_{\mb}} \times \intraencoderweightmatrixdimension}$; ${\weightm}_{{\lh}{'}}^{(n)}$, ${\weightm}_{\lh}^{(n)}$ and ${\weightm}_{m}^{(n)}$ are the projection matrix with the size of $\intraencoderweightmatrixdimension \times \intraencoderweightmatrixdimension$; 
${\query}_{\mb}^{(n)} \in \mathbb{R} ^{{ {\wholebodycordinatenumber}_{\mb}\times {\intraencoderweightmatrixdimension} }}$, 
${\key}_{\lh}^{(n)}\in \mathbb{R} ^{{ ({\wholebodycordinatenumber_{\lh}+3})\times {\intraencoderweightmatrixdimension} }}$, 
and ${\valu}_{\lh}^{(n)}\in \mathbb{R} ^{{ ({\wholebodycordinatenumber_{\lh}+3})\times {\intraencoderweightmatrixdimension} }}$ are the query, key and value features respectively; 
${\amap}_{att,{\lh}{\mb}}^{(n)} \in \mathbb{R} ^{{ {\wholebodycordinatenumber}_{\mb}\times (\wholebodycordinatenumber}_{\lh}+3) }$ is the attention map calculated by the $\operatorname{Softmax}(\cdot)$. 
$\operatorname{FFN}(\cdot)$ is composed of multi-layer perceptrons (MLPs).

For the major body, the semantic relevance with the left hand could be leveraged to fuse the semantic interaction context into itself progressively. Similarly, we can also obtain the cross-context semantic interaction of the right hand for major body ${\Fusionoutput}_{{\rh}{\mb}}^{(last)} \in \mathbb{R} ^{{{\wholebodycordinatenumber}_{\mb}} \times \intraencoderweightmatrixdimension}$. Combing the above semantic-related features with the
ego features, we can obtain the expressive features ${\Fusionoutput}_{\mb} \in \mathbb{R} ^{{ {\wholebodycordinatenumber}_{\mb}\times 3{\intraencoderweightmatrixdimension} }}$:
\begin{align}
    & {\Fusionoutput}_{\mb} = \operatorname{Concat}({\Fusionoutput}_{{\rh}{\mb}}^{(last)};{\Fusionoutput}_{{\lh}{\mb}}^{(last)};{\intrafeatures}_{m}),
\end{align}

\noindent where $\operatorname{Concat(\cdot)}$ is the concatenate operation along the feature dimension.
We also attain the features ${\Fusionoutput}_{\lh}\in \mathbb{R} ^{{ {(\wholebodycordinatenumber}_{\lh}+3)\times 3{\intraencoderweightmatrixdimension} }}$,
${\Fusionoutput}_{\rh}\in \mathbb{R} ^{{ ({\wholebodycordinatenumber}_{\rh}+3)\times 3{\intraencoderweightmatrixdimension} }}$ for left and right hand.

{\textbf{Physical Interaction.}} 
As the bridge between the body and hands, the wrist offers direct chain correlation between these two components. Therefore, we apply the `divide-and-fusion' strategy. That is to say, we first replicate the wrist joint to involve it, as illustrated in section~\ref{subsec:intra} for body and hand independently, and then perform dynamic feature fusion to form the final wrist features.
In this way, the physical connection between body parts could be better modeled. 
Specifically, we identify the feature of the wrist in $\{$major body, left hand$\}$ as the complementary pair. 
It is fed into MLPs to measure the mutual confidence, which are used as a weight to fuse paired features for more informed inference:
\begin{align}
    &{\lhMODoutput}        = {\lhmbMODweights} {\lhMODinput} + (1 - {\lhmbMODweights}) {\mbMODinput},                       \\
    &{\lhmbMODweights}  = \frac{1}{1 + \exp\left(  -{\tempweight} * {\ModeratorLayer}( {\lhMODinput}, {\mbMODinput})  \right)},      
\end{align}

\noindent where ${\lhMODinput} \in \mathbb{R} ^{3 \times 3\intraencoderweightmatrixdimension }$        
(or ${\mbMODinput} \in \mathbb{R} ^{3 \times 3\intraencoderweightmatrixdimension }$ )  is the features of the wrist in the left hand (or major body); and ${\lhmbMODweights}$ is the importance weight. ${\lhMODoutput}\in \mathbb{R} ^{3 \times 3\intraencoderweightmatrixdimension }$ is the fused wrist features. {${\tempweight}$ is a learnable temperature coefficient, jointly trained with all network parameters.} 
Similarly, we obtain the fused wrist features ${\rhMODoutput}\in \mathbb{R} ^{3 \times 3\intraencoderweightmatrixdimension }$ for $\{$major body, right hand$\}$.

\begin{table*}[t!]
\renewcommand\arraystretch{0.95}% 行距
\small
\centering
\setlength{\tabcolsep}{1.1mm}{
\begin{tabular}{cc|ccc|ccc|ccc|ccc|ccc|ccc}
\hline
\multicolumn{2}{c|}{Body Parts} &\multicolumn{3}{c|}{Major body} &\multicolumn{3}{c|}{Left Hands} & \multicolumn{3}{c|}{Right Hands} &\multicolumn{3}{c|}{Left Hands (AW)} &\multicolumn{3}{c|}{Right Hands (AW)} &\multicolumn{3}{c}{Whole Body}\\
% \hline
\multicolumn{2}{c|}{Time (sec)} & 0.2 & 0.4 & 1.0    & 0.2 & 0.4 & 1.0    & 0.2 & 0.4  & 1.0   & 0.2 & 0.4  & 1.0 & 0.2 & 0.4  & 1.0   & 0.2 & 0.4 & 1.0       \\
\hline
& LTD (D)
&8.7  &18.9  &48.7
&19.7 &57.0  &181.5
&33.3 &77.5 &195.6
&9.1  &18.3   &41.4
&17.2 &28.3   &53.1
&18.3 &45.6  &126.1
\\
& DMGNN (D)
&11.2  &23.1     &53.5     
&24.8  &62.0   &190.1    
&38.1  &83.0    &205.7 
&10.0  &21.7    &44.4
&21.6  &32.6    &60.5
&23.0  &55.7   &131.4
\\
&PGBIG (D)
&10.4  &21.7  &52.8     
&22.8  &61.5    &186.7    
&37.6  &82.4  &203.9    
&10.5  &22.2   &43.5
&21.5  &31.1  &58.7
&22.6  &53.6  &129.6
\\
% 右手的结果需要更新
\multirow{-4}{*}{\rotatebox[origin=c]{90}{\cellcolor{white}\textbf{Divided}}}
& {SPGSN (D)}
&9.3  &21.0 &52.6 
&25.3 &61.1 &164.2
&37.2 &81.5 &202.8
&9.3  &18.5  &41.6 
&16.1 &28.8   &56.9
&21.2 &48.4 &124.0
\\

\hline
\hline
% RH 结果有点问题，需要重新训练
{\cellcolor{white}} & {\cellcolor{white}LTD (U)}
&9.1 &19.9  &50.2
&19.9 &50.5  &162.5
&32.3 &74.6 &195.5
&8.9 &17.1  &42.5
&16.7 &29.3 &58.3
&18.4 &43.1 &120.4
\\
% &Res. sup.~\cite{martinez2017human}& \\
{\cellcolor{white}} & DMGNN (U) 
&13.7  &26.4    &56.9     
&22.4  &57.3   &172.0    
&36.3  &78.9   &203.7 
&9.7  &20.3   &46.4
&19.0  &33.2   &64.1
&22.7  &50.0  &128.2
\\
{\cellcolor{white}} &{PGBIG (U)} 
&13.2  &24.9   &54.2      
&23.0  &56.4   &165.7     
&35.0  &77.2   &199.4     
&10.2  &19.5   & 45.7
&19.1  &32.5  &62.0
&22.2  &48.1   &125.8
\\
{\cellcolor{white}} & {SPGSN (U)}   
&12.7 &24.5  &53.4
&21.6 &55.5 &161.6
&34.3 &75.5  &190.8
&9.6 &18.2 &42.3
&18.5 &31.0 &58.2
&21.0 &46.9  &120.3
\\
\hline
\hline

\multirow{-7}{*}{\rotatebox[origin=c]{90}{\cellcolor{white}\textbf{United}}}

& {\cellcolor{white}\textbf{EAI} (Ours)}
&\textbf{8.3} &\textbf{18.7} &\textbf{46.8}
&\textbf{17.7} &\textbf{49.2}  &\textbf{136.4}
&\textbf{29.8} &\textbf{69.0}  &\textbf{169.0}
&\textbf{8.6} &\textbf{17.3} &\textbf{38.8}
&\textbf{16.2} &\textbf{27.8} &\textbf{51.6}
&\textbf{16.7} &\textbf{40.7} &\textbf{104.6}
\\
\hline

\end{tabular}
}
\vspace{-.8em}
\caption{Average results on all actions with the evaluation metrics  MPJPE and MPJPE-AW (in $mm$). (AW), (D) and (U) are the abbreviation of the MPJPE-AW, divided and united training strategies. A lower value means better performance. The best results are highlighted in bold. Notably, {EAI} is only trained with united training strategies due to the need to explore body components' interactions.
(1) Compared with other baselines using a divided strategy, which lacks the interaction of components, our EAI outperforms them on all actions. It indicates the necessity of cross-context interaction. (2) Regarding the united strategy, which does not consider the heterogeneity of different body parts, our results are superior to other methods, which reveals the effectiveness of our cross-context alignment.}
\label{tab:united}
\vspace{-.2em}
\end{table*}

Then, the final expressive features are further reorganized as follows:
\textbf{(1)} Removing the left (right) wrist features ${\Fusionoutput}_{{\lh}}^{lw}$/${\Fusionoutput}_{{\rh}}^{rw}$ from the ${\Fusionoutput}_{\lh}$/${\Fusionoutput}_{\rh}$, the final features of left/right hand changes to $\widetilde{\Fusionoutput}_{\lh} \in \mathbb{R} ^{{ {\wholebodycordinatenumber}_{\lh} \times 3{\intraencoderweightmatrixdimension} }}$ ($\widetilde{\Fusionoutput}_{\rh} \in \mathbb{R} ^{{ {\wholebodycordinatenumber}_{\rh} \times 3{\intraencoderweightmatrixdimension} }}$);
\textbf{(2)} As to the body, after physical wrist refinement, the dimension of the feature is unchanged, but with 
${\Fusionoutput}_{{\mb}}^{lw}$/${\Fusionoutput}_{{\mb}}^{rw}$ updated by ${\lhMODoutput}$/${\rhMODoutput}$ to generate the final features $\widetilde{\Fusionoutput}_{\mb} \in \mathbb{R} ^{{ {\wholebodycordinatenumber}_{\mb} \times 3{\intraencoderweightmatrixdimension} }}$; \textbf{(3)} The resulting features ${ \{\widetilde{\Fusionoutput}_{\lh}}, {\widetilde{\Fusionoutput}_{\mb}}, {\widetilde{\Fusionoutput}_{\rh}}\}$ are then followed by a predictor composed of a MLP and IDCT to regress the final features into predicted sequence ${\{\mbpfposes},\lhpfposes, \rhpfposes\}$, where
${\lhpfposes} \in \mathbb{R} ^{{ {\wholebodycordinatenumber}_{\lh} \times \futuretime }}$,
${\mbpfposes} \in \mathbb{R} ^{{ {\wholebodycordinatenumber}_{\mb} \times \futuretime}}$,
 and
${\rhpfposes} \in \mathbb{R} ^{{ {\wholebodycordinatenumber}_{\rh} \times \futuretime }}$.

\textbf{Training Loss:} Prediction Loss $\mathcal L^{p}_{l}$ is defined to to measure the accuracy of the predicted 3D coordinates, we calculate the mean per joint position error:
\begin{equation}
    \resizebox{0.7\linewidth}{!}{$ \mathcal L^{p}_{l} = \frac{1}{N_{l}{\futuretime}}\sum_{n=1}^{N_{l}}\sum_{t=1}^{\futuretime}\|\hat{\textbf{x}}_{n,t}-\textbf{x}_{n,t}\|\;,$}
\end{equation}
\noindent where $\hat{\textbf{x}}_{n,t}\in \mathbb{R}^{3}$ denotes the predicted $n$-th joint position in frame $t$, ${\textbf x}_{n,t}$ the corresponding ground truth (GT).
$N_{l}$ the number of joints in the left hand skeleton. Similarly, we can also achieve $\mathcal L^{p}_{r}$ and $\mathcal L^{p}_{m}$ for right hand and body, forming the prediction loss of whole body $ \mathcal L^{p}=\mathcal L^{p}_{l}+\mathcal L^{p}_{m}+\mathcal L^{p}_{r}$.

Additionally, to further consider the hand semantics, we preprocess the gestures to be aligned with its wrist:

\begin{equation}
    \resizebox{0.7\linewidth}{!}{$ \mathcal L^{pw}_{l} = \frac{1}{N_{l}{\futuretime}}\sum_{n=1}^{N_{l}}\sum_{t=1}^{\futuretime}\|\hat{\textbf{x}}_{n,t}^{w}-\textbf{x}_{n,t}^{w}\|\;,$}
\end{equation}
\noindent where $\hat{\textbf{x}}^{w}_{n,t}\in \mathbb{R}^{3}$ denotes the predicted $n$-th joint position aligned with the left wrist, ${\bf x}_{n,t}^{w}$ is the corresponding GT. Then, we can obtain the fine-grained prediction loss of two hands $ \mathcal L^{pw}=\mathcal L^{pw}_{l}+\mathcal L^{p}_{r}$.

Since bone length is fixed for a human skeleton, we introduce the bone length loss:
\begin{equation}
    \resizebox{0.7\linewidth}{!}{$ \mathcal L^{b}_{l} = \frac{1}{(N_{l}-1){\futuretime}}\sum_{n=1}^{(N_{l}-1)}\sum_{t=1}^{\futuretime}|\hat{{b}}_{n,t}-{b}_{n}|\;,$}
\end{equation}
\noindent where $\hat{{b}}_{n,t}$ denotes the length of $n$-th bone, and ${b}_{n}$ the GT.  
$ \mathcal L^{b}=\mathcal L^{b}_{l}+\mathcal L^{b}_{m}+\mathcal L^{b}_{r}$ the bone length loss of whole body.

 To alleviate the features heterogeneity of different body components, we utilize the minimum distribution discrepancy error, proposed in section~\ref{subsec:XCA}, as the alignment Loss
\begin{align}
&\mathcal L^{a} = \mathcal L^{dis}_{lm} + \mathcal L^{dis}_{mr} + \mathcal L^{dis}_{rl}
\end{align}

{\textbf{Final Loss,}} is the weighted sum of the above losses:
\begin{equation}
\begin{aligned}
 \mathcal L = {\lambda_{1}}{\mathcal L^{p}} + {\lambda_{2}}({\mathcal L^{pw}} +{\mathcal L^{b}})+{\lambda_{3}} {\mathcal L^{a}},
\end{aligned}
\end{equation}

\noindent where $\lambda_1$, $\lambda_2$, $\lambda_3$ are the trade-off parameters. 

\section{Experiments} 

\label{sec:experiments}
\newcommand{\tb}{\textbf}

%%%%%%%%%%%%%%%%%%%%%%%%%%%%%%%%%%%%%%%%%%%%%%%%%%%%%%%%%%%%%%%%%%%%%%%
\begin{table*}[t!]
\renewcommand\arraystretch{0.95}%行距
\centering
\small
\setlength{\tabcolsep}{1.05mm}{
\begin{tabular}{cc|ccc|ccc|ccc|ccc|ccc|ccc}
\hline
\multicolumn{2}{c|}{Action} 
& \multicolumn{3}{c|}{A1 pass } 
& \multicolumn{3}{c|}{A2 eat } 
& \multicolumn{3}{c|}{A3 drink } 
& \multicolumn{3}{c|}{A4 lift } 
& \multicolumn{3}{c|}{A5 on } 
& \multicolumn{3}{c}{A6 squeeze }  \\
\multicolumn{2}{c|}{Time (sec)} 
& 0.2 & 0.4  & 1.0 
& 0.2 & 0.4 & 1.0 
& 0.2 & 0.4  & 1.0 
& 0.2 & 0.4  & 1.0 
& 0.2 & 0.4  & 1.0 
& 0.2 & 0.4  & 1.0 
\\
\hline

%%%%%%%%%%%%. main body %%%%%%%%%%%%%%%%%%%%
{\cellcolor{white}} & {\cellcolor{white}LTD (U)}
&10.2 &20.8  &42.4
&12.1 &28.0  &\textbf{71.7}
&12.2 &23.4 &40.1
&7.9 &20.3  &54.9
&9.8 &17.8  &36.4
&5.6 &12.7  &26.6
\\
{\cellcolor{white}} & {\cellcolor{white}DMGNN (U)} 
&11.7  &26.4    &40.7  
&17.9  &37.6    &88.4  
&14.5  &32.1    &58.0
&12.1  &26.3    &61.5  
&12.2  &22.0    &42.5  
&23.1  &48.1    &75.4  
\\
{\cellcolor{white}} &{\cellcolor{white}PGBIG (U)} 
&12.0  &26.9    &38.9 
&17.5  &36.5    &83.2  
&15.7  &30.2    &53.2 
&11.4  &24.3    & 62.4 
&13.0  &20.7    &41.2 
&21.5  &46.2    & 72.4 
\\
{\cellcolor{white}} & {\cellcolor{white}SPGSN (U)} 
&13.1 &25.8  &35.1
&18.4 &34.8  &82.0
&15.6 &28.9  &48.6
&10.6 &22.6  &51.6
&12.4 &21.7  &39.3
&7.9 &13.8  &27.9
\\

\multirow{-5}{*}{\rotatebox[origin=c]{90}{\cellcolor{white}\textbf{Major body}}}
{\cellcolor{white}} &{\cellcolor{white} \textbf{EAI} (Ours)}
   &\textbf{9.0} &\textbf{19.7}  &\textbf{31.6}
   &\textbf{10.5} &\textbf{26.4}  &{72.5}
   &\textbf{10.2} &\textbf{19.1}  &\textbf{30.8}
   &\textbf{6.5} &\textbf{16.4} &\textbf{44.4}
   &\textbf{7.9} &\textbf{15.6}  &\textbf{31.3}
   &\textbf{5.4} &\textbf{12.2}  &\textbf{24.4}
\\
\hline
\hline

%%%%%%%%%%%%. left hand %%%%%%%%%%%%%%%%%%%%
{\cellcolor{white}} & {\cellcolor{white}LTD (U)}
&\textbf{24.4} &\textbf{52.2}  &211.2
&21.7 &52.5   &187.5
&51.8 &123.4  &185.8
&21.0 &66.3   &163.8
&12.1 &34.1    &50.1
&18.1 &35.6    &54.6
\\
{\cellcolor{white}} & {\cellcolor{white}DMGNN (U)} 
&36.7  &68.9    &196.6 
&38.6  &87.5    &234.4  
&56.2  &128.8    &265.4  
&22.2  &68.7    &181.2  
&15.3  &44.3   &54.2  
&24.1  &49.3   &73.1 
\\
{\cellcolor{white}} &{\cellcolor{white}PGBIG(U)}
&33.5  &66.3   &186.2 
&36.9  &88.2    & 225.6
&56.7  &126.4   & 264.6
&23.1  &66.9    & 178.4 
&15.0  &43.2    & 50.1 
&23.0  &45.6    & 72.4 
\\
{\cellcolor{white}} & {\cellcolor{white}SPGSN (U)} 
&30.9 &71.1   &165.1
&36.5 &94.6   &263.6
&51.4 &119.8  &242.7
&20.3 &65.2   &175.5
&14.9 &41.1   &53.7
&22.9 &47.3   &70.3
\\
\multirow{-5}{*}{\rotatebox[origin=c]{90}{\cellcolor{white}\textbf{Left hands}}}
{\cellcolor{white}} &{\cellcolor{white} \textbf{EAI} (Ours)}

   &{25.4} &{52.6}  &\textbf{145.1}
   &\textbf{17.8} &\textbf{49.0}  &\textbf{148.7}
   &\textbf{42.5} &\textbf{107.7} &\textbf{144.4}
   &\textbf{14.3} &\textbf{48.4}  &\textbf{129.7}
   &\textbf{9.6} &\textbf{28.5}  &\textbf{45.8}
   &\textbf{10.8} &\textbf{32.9}  &\textbf{42.2}
\\
\hline
\hline

%%%%%%%%%%%%%%%%% right hand %%%%%%%%%%%%%%%%
{\cellcolor{white}} & {\cellcolor{white}LTD (U)}
&37.0 &82.1  &136.1
&35.3 &79.3  &204.3
&{22.9} &82.2 &167.2
&25.5 &81.5  &229.1
&45.1 &97.0  &187.2
&25.1 &47.8 &93.7
\\
% &Res. sup.~\cite{martinez2017human}& \\
{\cellcolor{white}} & {\cellcolor{white}DMGNN (U)} 
&39.2  &80.5   &129.5  
&37.5  &78.3    &215.0 
&23.5  &85.8   &221.4
&27.3  &83.4    &231.0  
&47.3  &105.6  &230.2  
&26.4  &54.2   &103.7  
\\
{\cellcolor{white}} &{\cellcolor{white}PGBIG (U)} 
&36.8  &78.3   &124.6
&34.2  &76.4  &212.5  
&24.0  &87.6   &210.5 
&26.1  &82.5   &233.7 
&47.2  &103.4   &221.9
&25.7  &54.0    &102.5 
\\
{\cellcolor{white}} & {\cellcolor{white}SPGSN (U)} 
&33.7 &73.0  &108.7
&{31.8} &\textbf{59.5} &207.6
&22.5 &92.0  &249.4
&21.3 &76.4 &215.6
&42.4 &101.4 &173.5
&24.7 &52.6  &98.7
\\
\multirow{-5}{*}{\rotatebox[origin=c]{90}{\cellcolor{white}\textbf{Right hands}}}
{\cellcolor{white}} &{\cellcolor{white} \textbf{EAI} (Ours)}

&\textbf{21.7} &\textbf{50.3} &\textbf{69.6}
&\textbf{31.8} & {70.2}  &\textbf{180.3}
&\textbf{15.2} &\textbf{60.8}  &\textbf{111.0}
&\textbf{17.6} &\textbf{51.0}  &\textbf{136.9}
&\textbf{35.1} &\textbf{79.5}  &\textbf{146.2}
&\textbf{23.1} &\textbf{46.1}  &\textbf{86.6}
\\
\hline
\end{tabular}}
\vspace{-.8em}
\caption{Detailed results on common action split with the evaluation metrics MPJPE (in $mm$). (U) is the abbreviation of united training strategy. The best results are highlighted in bold. We observe that for both fine- and coarse-grained motion patterns, our results consistently outperform the competitors. 
It evidences the compatibility of the EAI for various activities.
}
\label{tab:wholebody}
\vspace{-1.7em}
\end{table*}

\textbf{Datasets:}
\label{sec:experimentsdatasetes}
To our knowledge, previous widely-used datasets, \textit{e.g.,} H3.6M \cite{ionescu2013human3}, 3DPW \cite{vonMarcard2018}, only record the major body motions (without human hands). To be compatible with our proposed novel task, here we select the GRAB~\cite{taheri2020grab}.
It is a recently released dataset with over 1.6 million frames of 10 different actors performing a total of 29 actions.

It is captured using high-precision motion capture techniques.
GRAB provides SMPL-X \cite{pavlakos2019expressive} parameters from which we extract 25 joints (3D position) defined as the body ($N_m=25$), and each hand is represented as 15-joints ($N_l = N_r = 15$). 

% Since delicate hand movements are more common in indoor activity and the GRAB dataset contains a vast array of movements, which suffices to complete our goal of human-robot interaction. As a result, no additional datasets were utilized for training. More details can be found in the \textbf{supplementary materials}.

\textbf{Baselines:}
\label{subsec:baseline}
We note that for 3D whole-body human motions forecasting, there are no direct methods for comparisons.
Therefore, to comprehensively investigate the proposed EAI, we select 4 SOTA approaches of standard major-joint prediction as our baselines, including LTD \cite{mao2019learning}, DMGNN \cite{li2020dynamic}, PGBIG \cite{ma2022progressively}, SPGSN  \cite{li2022skeleton}.
Notably, all baselines are based on GCNs to consider the $N$-joint human skeleton ($N=17$ or $N=25$). 
To a fair comparison, we retrain the baselines under the following training setups.

We apply two training strategies to investigate this new task.
\textbf{(1)} For the \textbf{divided (D)} training, we separately train the baselines for each human components. 
This independent strategy lacks the interaction of components and thus can be used to illustrate the effectiveness of XCI. 
\textbf{(2)} For the \textbf{united (D)} training, we extend the node number of GCNs to 55 ($N_m=25$, $N_l=N_r=15$), as in our experimental setup. This strategy implicitly contains the cross-context interaction via a whole-body graph but does not consider the heterogeneity of different body parts. 
Therefore, it is used to demonstrate the effectiveness of XCA. 

\textbf{Training details:} We employ AdamW \cite{loshchilov2017decoupled} optimizer with an initial learning rate of 0.001 and batch size of 64 to train our model (50 epochs).
The learning rate is decayed by 0.96 for every two epochs. 
The trade-off parameters {$\{\lambda_1,\lambda_2,\lambda_3\}$} are set as $\{ 1,0.1,0.001\}$. 
More details are set in the \textbf{supplementary materials}. 

\begin{figure*}[t]
\begin{center}
  \includegraphics[width=0.9 \textwidth]{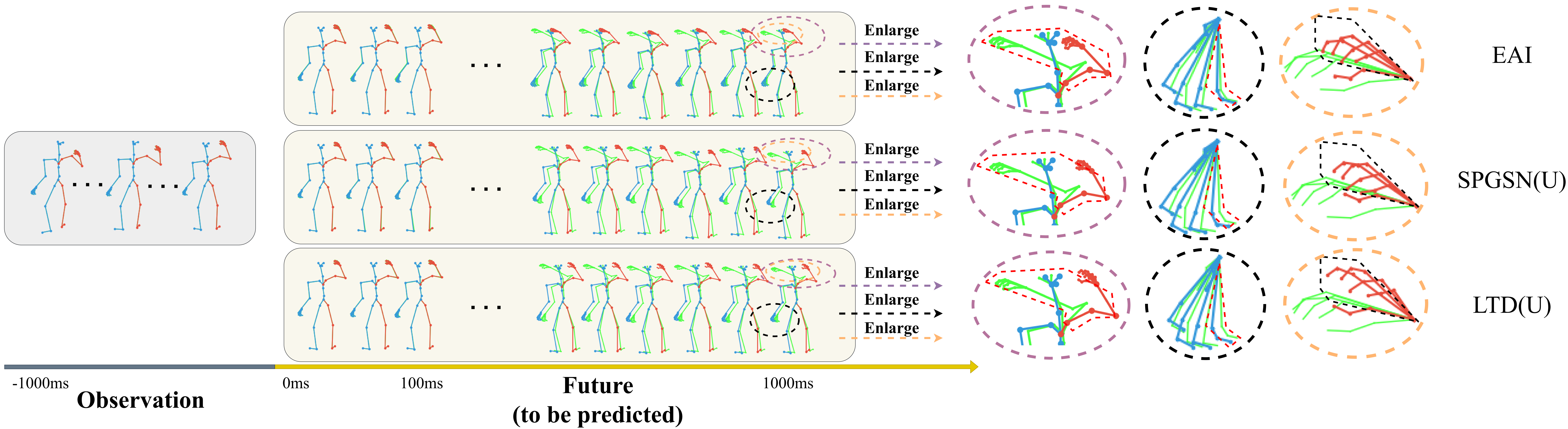} 
\end{center}
\vspace{-1em}
\caption{Predicted whole-body poses visualization (skeleton).
The past sequence is in a grey box, and the predicted ones are in yellow boxes.
The GT and predicted poses are denoted as green and blue/red skeletons, respectively. 
As highlighted by the dashed ellipse boxes, both performances of fine-grained (body) and coarse-grained (gestures) motion are considered.
This evidences that it is indeed beneficial to simultaneously eliminate the heterogeneity of different human components and then extract the interaction within the whole-body.
}
\vspace{-1em}
\label{Fig:vis_whole}
\end{figure*}

\textbf{Metrics:}
\label{sec:experimentsmetrics}
For the whole-body motion, we use the mean per joint position error (MPJPE) \cite{mao2019learning,mao2020history,li2020dynamic,ma2022progressively} to measure the 3D prediction accuracy of overall movement.
Besides, since there are no baselines for hand prediction, we extend the baseline of major body motion prediction~\cite{mao2019learning,mao2020history,li2020dynamic,ma2022progressively} into hand prediction and also leverage MPJPE to measure the prediction accuracy. However, the MPJPE of hands is affected by wrist movement severely, which is not able to show subtle hand activities and semantic information.
Therefore, we also report the MPJPE-AW after alignment\cite{martinez2017simple} with the wrist.

\subsection{Comparison with the SOTA methods} 
\label{subsec:exp}

{\textbf{Baselines (U) v.s. Baselines (D).}}
Table~\ref{tab:united} shows the average prediction error of all actions between our method and the above four baselines. The baselines are trained with two strategies: divided and united, as illustrated in section~\ref{subsec:baseline}. Notably, because the MPJPE of hands is severely affected by wrist movement, we also show the MPJPE-AW on the prediction of delicate hand movement. Compared with the divided strategy, the predicted results for the body are worse when training unitedly. And the hands' results show opposite trends on the two metrics. 
The above result reveals that:
\textbf{(1)} 
The interaction is indeed meaningful to improve prediction accuracy (MPJPE of hands is lower). 
\textbf{(2)} 
However, the implicit modeling of interaction within a whole-body graph may bring negative mutual interference (MPJPE of body and MPJPE-AW of hands are higher) because major body and gestures have heterogeneous motion patterns.

{\textbf{EAI v.s. Baselines (U$\&$D).}}
Our proposed EAI addresses the above two limitations of existing methods. 
\textbf{(1)} As to the united strategy, EAI is superior to all baselines by a large margin. It verifies the effectiveness of cross-context alignment (XCA), which considers the motion heterogeneity of different body parts.  
\textbf{(2)} Compared with the baseline results using the divided strategy, our method is better, which demonstrates that cross-context interaction (XCI) across body components is vital.
Both rough (major joints) and delicate (gestures) properties are cross-facilitated to achieve a higher-fidelity prediction via the EAI framework.

{\textbf{Compatibility. }}
Table~\ref{tab:wholebody} shows more detailed results on common action with the evaluation metrics MPJPE. The error obtained by our method is smaller than others in most cases. 
The activity with both fine-grained (drink $\&$ eat) and coarse-grained (lift $\&$ pass) motion patterns achieve more improvements than the baseline approaches, which evidences the compatibility of our proposed EAI. 
Moreover, the enhanced performance on all body parts also verifies the necessity of considering both the heterogeneity and interactivity across different body parts.
Results of the other actions can be found in the \textbf{supplementary material}.

{\textbf{Visualization.}}
In Figure~\ref{Fig:vis_whole}, we show the whole-body qualitative results of the 'play' action via the skeletal form. 
As highlighted by the purple dashed ellipse, the absolute prediction of upper limbs and hands is much closer to ground truth (denoted by solid green lines). It demonstrates that expressive context information extracted from EAI leads to the overall refinement of coarse-fined motion. 
Besides, the other two dashed ellipses show fine-grained gestures by aligning the hand sequence with the wrist. We observe that EAI still outperforms other baselines in the relative results, which illustrates that the delicate semantic information of gestures could be better considered.  
The results of fine-grained and coarse-grained motion are enhanced, verifying the significance of co-analyzing  different body components for the novel whole-body pose forecasting task.

\subsection{Ablation Studies} 
We conduct ablation studies on model architecture for deeper analysis.
More discussions are in the \textbf{supplementary materials}.
We run experiments under the condition of separately removing the XCA and XCI, as well as the following sub-modules:
(a) cross neutralization (CN), (b) discrepancy constraint (DC) in XCA; (c) semantic interaction (SI), (d) physical interaction (PI) in XCI.

Table \ref{tab:ablation} reports the detailed results. 
The full model contains both XCA and XCI, and the average prediction error is 61.9mm. (1) Without the CN and DC, the prediction error is 66.6mm, which is a noticeable performance drop, demonstrating the necessity to alleviate distribution discrepancy.
Removing CN/DC, the average error increases by 2.5/1.4mm. It shows the CN is more critical in XCA;
(2) Excluding the entire XCI, the prediction error drastically increases from 61.9mm to 68.7mm. 
This gap is larger than the case without the whole XCA, indicating that the interaction extraction is more vital than heterogeneity reduction relatively. Remarkably, the prediction error of XCI (\textit{w/o} SI) / XCI (\textit{w/o} PI) increased by 5.6/2.6mm. 
It reveals that the semantic relevance of body components is more valuable to perceive motion properties.

\begin{table}[!ht]
\small
\renewcommand\arraystretch{1.2}% 行距
\centering
\setlength{\tabcolsep}{1.65mm}{
\begin{tabular}{c|cccc|ccc|c}
\hline
 {} &{CN}&{DC} & {PI} & {SI}    &0.2s  & 0.4s   & 1.0s  & Avg.
\\ 
\hline
\cellcolor{white} &{} & \checkmark & \checkmark & \checkmark  &\textbf{16.7} &\textbf{40.7} &90.4 & 64.4   
\\
\cellcolor{white}&\checkmark &  & \checkmark & \checkmark  &\underline{17.0} &41.3 &\underline{87.9}  & \underline{63.3}  
\\
\multirow{-3}{*}{{\cellcolor{white}\textbf{XCA}}}
{\cellcolor{white}} 
& &  & \checkmark & \checkmark  &\underline{17.0} &42.8 &93.7  & 66.6  \\ 
            
\hline
\cellcolor{white}& \checkmark & \checkmark &  & \checkmark  &\textbf{16.7} &\underline{41.1} &89.8   & 64.5  \\
\cellcolor{white}&\checkmark & \checkmark &\checkmark  &     &\underline{17.0} &42.5 &94.2  & 67.5\\
\multirow{-3}{*}{{\cellcolor{white}\textbf{XCI}}}
&\checkmark & \checkmark &  &    &17.8 &43.2 &95.0  & 68.7 \\
\hline
{\textbf{Full model}}&\checkmark & \checkmark & \checkmark & \checkmark  & \textbf{16.7} & \textbf{40.7} & \textbf{85.8}  & \textbf{61.9} \\
\hline
\end{tabular}
}
\vspace{-.6em}
\caption{Ablation experiments of model architecture.
The best (second-best) result is highlighted in bold (underlined). }
\vspace{-.6em}
\label{tab:ablation}
\end{table}

\section{Conclusion} \label{sec:conclusion}
In this work, we introduce a new task: expressive forecasting of 3D whole-body human motions. 
To tackle this challenge, we propose a novel Encoding-Alignment-Interaction (EAI) framework that takes into account the heterogeneous information within the whole body and the collaboration among various human components.
Our approach jointly considers the heterogeneous information within the whole body and the interaction/collaboration among various human components.
Compared with conventional predictive algorithms, EAI could cross-facilitate both coarse- (body) and fine-grained (gestures) properties. 
Extensive experiments demonstrate that the proposed approach achieve the superior performance and surpasses the state-of-the-art methods by a large margin.
Considering the downstream application of whole-body forecasting, we conclude that the proposed model is of practical importance; however, there are areas that need further exploration in the future. 
For instance, incorporating interactions with objects could provide vital cues to improve the accuracy of motion anticipation.

\section{Acknowledgments}
This work was supported by the National Science and Technology Innovation 2030 - Major Project (Grant No. 2022ZD0208800), and NSFC General Program (Grant No. 62176215).
This work was supported in part by the National Natural Science Foundation of China (62306141), in part by the Jiangsu Funding Program for Excellent Postdoctoral Talent (2022ZB269), in part by the Natural Science Foundation of Jiangsu Province (BK20220939), and in part by the China Postdoctoral Science Foundation (2022M721629).

\bibliography{aaai24}

\end{document}